%% file: PaperForReview.tex
\documentclass[10pt,twocolumn,letterpaper]{article}

\usepackage[pagenumbers]{cvpr} % To force page numbers, e.g. for an arXiv version

\usepackage{graphicx}
\usepackage{amsmath}
\usepackage{amssymb}
\usepackage{booktabs}
\usepackage{multirow}

\usepackage[pagebackref,breaklinks,colorlinks]{hyperref}

\usepackage[capitalize]{cleveref}
\crefname{section}{Sec.}{Secs.}
\Crefname{section}{Section}{Sections}
\Crefname{table}{Table}{Tables}
\crefname{table}{Tab.}{Tabs.}

\def\cvprPaperID{*****} % *** Enter the CVPR Paper ID here
\def\confName{CVPR}
\def\confYear{2023}

\begin{document}

%\title{Facial Affective Analysis based on MAE and Multi-modal Information for 5th ABAW Competition}
\title{Multi-modal Facial Affective Analysis based on Masked Autoencoder}

% \and
\author{Wei Zhang, 
Bowen Ma, 
Feng Qiu,  
Yu Ding\footnotemark[2]\thanks{Corresponding Author.}\\
Virtual Human Group, Netease Fuxi AI Lab\\
{\tt\small \{zhangwei05,mabowen01,qiufeng,dingyu01\}@corp.netease.com}
}

\maketitle

%%%%%%%%%%%%%%%%%%%%%%%%%%%%%%%%%%%%%%%%%%%%%%%%%%%%%%
\input{0_Abstract}

%%%%%%%%%%%%%%%%%%%%%%%%%%%%%%%%%%%%%%%%%%%%%%%%%%%%%%
\input{1_Introduction}

%%%%%%%%%%%%%%%%%%%%%%%%%%%%%%%%%%%%%%%%%%%%%%%%%%%%%%
\input{2_RelatedWorks}

%%%%%%%%%%%%%%%%%%%%%%%%%%%%%%%%%%%%%%%%%%%%%%%%%%%%%%
\input{3_Method}
\input{4_Experiments}
\input{5_Conclusion}
{\small
\bibliographystyle{ieee_fullname}
\bibliography{egbib}
}

\end{document}

%% file: 0_Abstract.tex
\begin{abstract}
Human affective behavior analysis focuses on analyzing human expressions or other behaviors to enhance the understanding of human psychology. 
The CVPR 2023 Competition on Affective Behavior Analysis in-the-wild (ABAW) is dedicated to providing high-quality and large-scale Aff-wild2 for the recognition of commonly used emotion representations, such as Action Units (AU), basic expression categories~(EXPR), and Valence-Arousal (VA). The competition is committed to making significant strides in improving the accuracy and practicality of affective analysis research in real-world scenarios.
In this paper, we introduce our submission to the CVPR 2023: ABAW5. 
Our approach involves several key components. First, we utilize the visual information from a Masked Autoencoder~(MAE) model that has been pre-trained on a large-scale face image dataset in a self-supervised manner. 
Next, we finetune the MAE encoder on the image frames from the Aff-wild2 for AU, EXPR and VA tasks, which can be regarded as a static and uni-modal training. 
Additionally, we leverage the multi-modal and temporal information from the videos and implement a transformer-based framework to fuse the multi-modal features. Our approach achieves impressive results in the ABAW5 competition, with an average F1 score of 55.49\% and 41.21\% in the AU and EXPR tracks, respectively, and an average CCC of 0.6372 in the VA track. Our approach ranks first in the EXPR and AU tracks, and second in the VA track. Extensive quantitative experiments and ablation studies demonstrate the effectiveness of our proposed method. 
\end{abstract}

%% file: 1_Introduction.tex
%%%%%%%%%%%%%%%%%%%%%%%%%%%%%%%%
\section{Introduction}
\label{sec:intro}

In recent years, there has been a growing interest in the research of human affective behavior analysis due to its potential to provide a more accurate understanding of human emotions, which can be applied to design more friendly human-computer interaction. 
The commonly used human expression representations include Action Unit~(AU), basic expression categories~(EXPR), and Valence-Arousal~(VA). Specifically, AU is first proposed by Paul Ekman and Wallace Friesen in the 1970s~\cite{ekman1978facs}. It depicts the local regional movement of faces which can be used as the smallest unit to describe the expression. Basic expression categories divide expressions into a limited number of groups according to the emotion categories, e.g., happiness, sadness, etc. VA contains two continuous values Valence~(V) and Arousal~(A), which are ranged between -1 and 1. They can be used to describe the human emotional state. V represents the degree of positivity or negativity of emotion; A describes the level of intensity or activation of emotion.
% ERI typically comprises a sequence of values representing multiple emotional dimensions that reflect the intensity of an individual's emotional response to a specific stimulus.

The fifth Competition on Affective Behavior Analysis in-the-wild~(ABAW5)~\cite{kollias2023abaw} is organized to focus on handling the obstacles in the process of human affective behavior analysis. It makes great efforts to construct large-scale multi-modal video datasets Aff-wild~\cite{zafeiriou2017aff,kollias2019deep,kollias2019face} and Aff-wild2~\cite{kollias2019expression,kollias2020analysing,kollias2021affect,kollias2021analysing,kollias2021distribution,kollias2022abaw2}. The proposal of these datasets has greatly promoted the development of facial expression analysis in the wild and accelerated the practical implementation of related industries.
Aff-wild2 contains 598 videos and most of them have the three kinds of frame-wise annotated labels: AU, basic expression categories and VA. ABAW5 proposes three challenges of detecting these three kinds of expression representations. 

% Besides, ABAW5 builds up a Hume-Reaction dataset which consists of about 75 hours of video recordings, recorded via a webcam, in the subjects’ homes. Each video in it has been self-annotated by the subjects themselves for the ERI intensity of 7 emotional experiences.

In this paper, we introduce our submission to the ABAW5. First of all, we pre-train a Masked Autoencoder (MAE)~\cite{he2022masked,ma2022maeface} on a large-scale facial dataset in a self-supervised manner. Then, we choose the MAE encoder as our visual feature extractor to capture the visual features of the faces. Due to the extensive quantity of faces included in the dataset, the features extracted with the MAE encoder have strong generalization capabilities. We also finetune the MAE encoder for the specific tasks of AU detection, EXPR classification, and VA estimation. This training process only use the static and vision modal data.
To further exploit temporal and multi-modal information, we design a Temporal and Multi-modal Fusion~(TMF) that divides the videos into several short clips and performed clip-wise training on the downstream tasks mentioned above. 
During this process, we utilized the finetuned MAE encoder to extract visual features from each frame, while also incorporating pre-trained audio models such as Hubert~\cite{hubert}, Wav2vec2~\cite{baevski2020wav2vec}, and Vggish~\cite{vggish} to capture acoustic features.
The visual and acoustic features are concatenated and fed into a Transformer structure to capture temporal information for the downstream tasks. Additionally, we proposed several effective post-processing policies aimed at smoothing predictions and further enhancing model performance.

In sum, the contributions of this work are two-fold:
\begin{itemize}
    \item  Our approach employs a highly effective MAE feature extractor to capture visual features. The MAE model is pre-trained on a large-scale facial image dataset and exhibits remarkable generalization ability for diverse downstream face-related tasks.
    \item  Our approach employs a Temporal and Multi-modal Fusion (TMF) to leverage the benefits of temporal and multi-modal information. With a strategic selection of optimal vision and audio features, our approach further improves the performance of the model. 
    \item  In the ABAW5 competition, our approach ranks first in both AU and EXPR tracks and ranks second in VA track. The final test set score and quantitative experiments can prove the superiority of our method.  
\end{itemize}
% Moreover, we design a dual-branch structure that contains Basic Learning Branch~(BLB) and Collaboration Learning Branch~(CLB). BLB and CLB have the same structure and shared feature extractors. By randomly interpolating the logit space of BLB and CLB, the model can enrich the feature space by implicitly creating some potential samples, which further enhance the model generalization.

%% file: 2_RelatedWorks.tex
%%%%%%%%%%%%%%%%%%%%%%%%%%%%%%%%
\section{Related Works}

In this section, we introduce some recent works on the relevant tasks in the CVPR2023: ABAW5 competition, including AU detection, expression recognition and VA estimation in the wild. We also briefly introduce the self-supervised learning method in facial affective analysis. 

%%%%%%%%%%%%%%%%
\subsection{AU Detection}

For AU detection in the wild, there exist some challenges regarding the limited identity information and interference of diversity poses, illumination or occlusions. These disturbances restrain the model generation and cause the overfitting to the noise information.
Several studies propose the use of a multi-task framework to incorporate additional auxiliary information as regularization, which introduces extra label constraints. Specifically, Zhang et al. \cite{multi-task1} proposed a streaming model that simultaneously performs AU detection, expression recognition, and VA regression. Similarly, Jin et al. \cite{multi-task2} and Thinh et al. \cite{multi-task3} combine the tasks of AU detection with expression recognition. JAA-Net~\cite{shao2021jaa} performs landmarks detection and AU detection at the same time. 

Another effective approach to enhance the model generalization is to utilize the related tasks' pre-trained backbones. 
Jiang et al. \cite{AU_2} use IResnet100~\cite{iresnet} that pre-trains on Glint360K~\cite{an2022pfc} and some private commercial datasets before conducting the AU detection task in the ABAW3. 
Savchenko et al.~\cite{EXPR4} utilize the EfficientNet~\cite{tan2019efficientnet} that pre-trained on the face recognition task as the backbone.
Zhang et al.\cite{multi-task1,zhang2022} introduce the pre-trained expression embedding model as the backbone and win the first prizes in ABAW2 and ABAW3.

Multi-modal information is also involved in ABAW competitions. 
Zhang et al.~\cite{zhang2022} capture three modalities of information - vision, acoustic, and text - and fused them using a transformer decoder structure. 
Jin et al.~\cite{jin2021multi} extract the vision features from IResNet100 and the audio features from Mel Spectrogram. They also use the transformer structure for the fusion of multi-modal features.

%%%%%%%%%%%%%%%%
\subsection{Expression Recognition}

The goal of expression recognition is to classify an input image into one of the basic emotion classes, such as happiness or sadness.
Similar approaches to exploit the extra information regularization are mentioned in Sec.~2.1. Zhang et al.~\cite{multi-task1} utilize the prior expression embedding model and propose a multi-task framework. 
Phan et al.\cite{EXPR5} employ the pre-trained model RegNet\cite{radosavovic2020designing} as the backbone and add the Transformer~\cite{vaswani2017attention} structure to extract the temporal information. 
Kim \textit{et al.} \cite{EXPR6} use Swin transformer \cite{liu2021swin} as the backbone and exploit the extra auxiliary from the audio modal. 
Wang et al. \cite{wang2021multi} propose a semi-supervised framework to predict pseudo-labels for unlabeled data, which helps improve the model's generalization to some extent.
Xue \textit{et al.}~\cite{EXPR3} develop a CFC network that uses different branches to train the easy-distinguished and hard-distinguished emotion categories.  

%%%%%%%%%%%%%%%%
\subsection{VA Estimation}
For VA estimation, several studies~\cite{multi-task1,ABAW2_VA1,ABAW2_VA3,ABAW2_VA4} leverage the correlation between VA and AU or VA and EXP, proposing multi-task frameworks. This enables these methods to extract supplementary information from other tasks, particularly for data without VA labels but possessing AU or EXPR labels.
Moreover, many works~\cite{ABAW3_VA1,ABAW3_VA2,ABAW5_AU1,ABAW5_EXPR5,ABAW3_VA5,ABAW5_AU3} propose multi-modal frameworks, which leverage hidden features from vision, audio, or text. The Transformer structure is also commonly used in VA tasks for feature fusion. Several works~\cite{ABAW3_VA1,ABAW5_AU1,ABAW5_AU2,ABAW5_EXPR3} utilize it for this purpose.

%%%%%%%%%%%%%%%%
\subsection{Self-supervised Learning in Affective Analysis}
It has been pointed out~\cite{zafeiriou2017aff} that annotating the corresponding emotion/AU/VA labels from real-world facial images costs a large amount of time/labor, which limits the development of the affective analysis. It is a potential solution to make use of the self-supervised learning~(SSL) method to exploit more knowledge from the existing large-scale unlabelled data. There have been several works to develop SSL methods to enhance the accuracy in the area of affective analysis. 
Shu et al.\cite{shu2022revisiting} explore different strategies in the choice of positives and negatives to enforce the expression-related features and reduce the interference of other facial attributes. They improve the accuracy of expression recognition based on the contrastive SSL methods (e.g. SimCLR~\cite{chen2020simple}).
Ma et al.~\cite{ma2022maeface} pre-train the Masked Autoencoder~(MAE) structure on the large-scale face images and finetune it on the AU detection, which achieves the state-of-the-art performance on the BP4D~\cite{zhang2014bp4d} and DISFA~\cite{mavadati2013disfa}.
In the ABAW5 competition, Zhang et al.~\cite{ABAW5_AU1}, Liu et al.~\cite{ABAW5_AU2}, and Wang et al.~\cite{ABAW5_AU4} all employ the pre-trained MAE to extract vision features and secure a position among the top few performers.

%% file: 3_Method.tex
\begin{figure*}
    \centering
    \includegraphics[ width=1\linewidth]{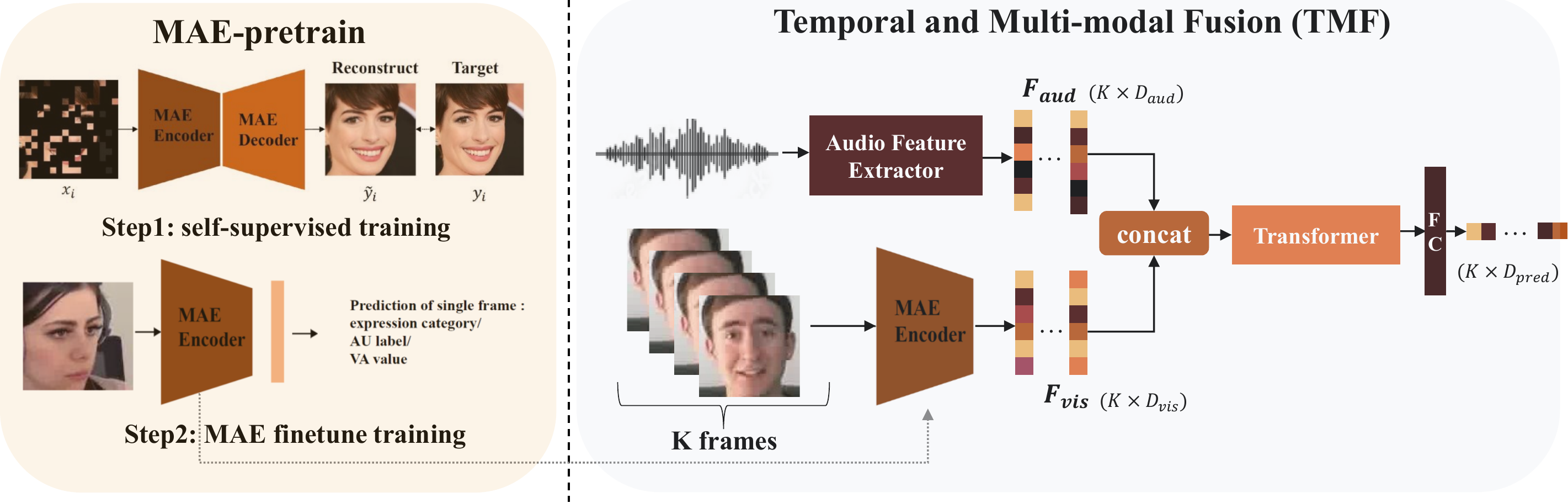}
    %\vspace{-1.5em}
    \caption{Illustration of our pipeline. In the first stage, we pre-train a MAE in  SSL manner. The input face image is randomly masked 75\% and MAE learns to reconstruct the missing pixels. Then, the MAE encoder is fine-tuned on the Aff-wild2 for AU detection, EXPR classification, and VA estimation. In the second stage, we propose a transformer-based structure TMF that utilizes temporal vision and audio information to further facilitate the challenge tasks. }
    \label{fig:pipeline}
\end{figure*}

%%%%%%%%%%%%%%%%%%%%%%%%%%%%%%%%
\section{Method}
This section will present our method proposed for AU detection, expression recognition and VA estimation in ABAW5 competition. The pipeline can be seen in Fig.~\ref{fig:pipeline}. It consists of two steps: MAE pre-train and Temporal and Multi-modal Fusion~(TMF). We first introduce the MAE structure with good generalization to extract the vision features. Subsequently, we employ the TMF to integrate the temporal acoustic features with the temporal visual features to extract more effective information for completing the task. Finally, we adopt some smoothing policies to generate the final predictions. 

%%%%%%%%%%%%%%%%
\subsection{MAE Pre-training}
Different from the traditional MAE, our MAE is pre-trained on the facial image dataset to focus on learning the facial vision features. We construct a large-scale facial image dataset that contains images from the existing facial image datasets, e.g., AffectNet~\cite{AffectNet}, CASIA-WebFace~\cite{CASIA-Webface}, CelebA~\cite{CelebA} and IMDB-WIKI~\cite{IMDB-WIKI}. The total number of images is about 2.17M.
Then we pre-train the MAE model on the dataset in a self-supervised manner. Specifically, our MAE consists of a ViT-Base encoder and a ViT decoder based on the structure of Vision Transformer (ViT)~\cite{ViT}. 
The MAE pre-training procedure follows a masking-then-reconstruct method, whereby images are first divided into a series of patches (16x16) and 75\% of them are randomly masked. 
These masked images are sent to the MAE encoder and the complete images should be reconstructed by the MAE decoder~(See Fig.~\ref{fig:pipeline} left Step1). The loss function of MAE pre-training is the pixel-wise L2 loss to make the reconstructed images close to the target images.

% \begin{figure}
%     \centering
%     \includegraphics[ width=1\linewidth]{images/MAE1.pdf}
%     %\vspace{-1.5em}
%     \caption{Description of MAE self-supervised pre-training and fine-tuning on the downstream tasks. (a) The input face image is randomly masked 75\% and MAE learns to reconstruct the missing pixels. (b) After SSL, the MAE encoder is fine-tuned on the Aff-wild2 for AU detection, EXPR classification, and VA estimation.}
%     \label{fig:mae}
% \end{figure}

Once the self-supervised learning is complete, we remove the MAE decoder and replace it with a fully connected layer attached to the MAE encoder (See Fig.~\ref{fig:pipeline} left Step2). This allows us to fine-tune downstream tasks: AU detection, expression recognition, and VA estimation on the Aff-wild2 dataset. It is important to note that this process is based on frame-wise training, without taking into account temporal or other modal information. The corresponding loss functions for these three tasks are as follows:

\begin{equation}
    \mathcal{L}_{\textit{AU\_CE}}=-\frac{1}{12}\sum_{j=1}^{12} W_{au_j}[y_{j}\log\hat{y}_{j} 
     + (1-y_{j})\log(1-\hat{y}_{j})]. 
\label{eq:1}
\end{equation}
\begin{equation}
    \mathcal{L}_{\textit{EXPR\_CE}}=-\frac{1}{8}\sum_{j=1}^{8}W_{exp_j} z_{j}\log\hat{z}_{j}.
\label{eq:2}
\end{equation}

\begin{equation}
\begin{split}
\mathcal{L}_{\textit{VA\_CCC}}= 1 - CCC(\hat{v}_{batch_i},v_{batch_i}) \\
     + 1 - CCC(\hat{a}_{batch_i},a_{batch_i}) 
\end{split}
\label{eq:3}
\end{equation}
\begin{equation}
\label{CCC}
    \textit{CCC}(\mathcal{X},\mathcal{\hat{X}})=\frac{2\rho_{\mathcal{X\hat{X}}}\delta_{\mathcal{X}}\delta_{\mathcal{\hat{X}}}}{\delta_{\mathcal{X}}^2+\delta_{\mathcal{\hat{X}}}^2+(\mu_{\mathcal{X}}-\mu_{\mathcal{\hat{X}}})^{2}}.
\end{equation}
where $\hat{y}$, $\hat{z}$, $\hat{v}$ and $\hat{a}$ denote the model's predictions for AU, expression category, Valence, and Arousal, respectively. The symbols without hats refer to the ground truth. $\delta_{\mathcal{X}}$,$\delta_{\mathcal{\hat{X}}}$ indicate the standard deviations of $\mathcal{X}$ and $\mathcal{\hat{X}}$, respectively. $\mu_{\mathcal{X}}$ and $\mu_{\mathcal{\hat{X}}}$ are the corresponding means and  $\rho_{\mathcal{X\hat{X}}}$ is the correlation coefficient.
For the AU and EXPR tasks, we utilize weighted cross-entropy as the loss function. The weights for different categories, represented by $W_{au_j}$ and $W_{exp_j}$, are inversely proportional to the class number in the training set. 

%%%%%%%%%%%%%%%%
\subsection{Temporal and Multi-modal Fusion}
To further exploit the temporal and multi-modal features for AU, EXPR and VA tasks, we design the sequence-based model which combines the audio features. 
To concretely, we first divide the videos into several short clips, each having an equivalent frame number of K. 
We construct the Temporal and Multi-modal Fusion~(TMF) to perform the sequence-wise training which can be seen in Fig.~\ref{fig:pipeline} right. 

Given a video clip $C_i$ and the corresponding audio clips $A_i$, we use the finetuned MAE encoder and some existing pre-trained audio feature extractor models~(e.g. Hubert~\cite{hubert}, Wav2vec2~\cite{baevski2020wav2vec}, vggish~\cite{vggish}.) to extract the vision and acoustic features $F^{i}_{vis}$ and $F^{i}_{aud}$ for each frame separately. 
Then we concatenate $F^{i}_{vis}$ and $F^{i}_{aud}$ and sent them into a Transformer~\cite{vaswani2017attention} encoder structure to exploit the temporal correlations between them. 
The Transformer encoder comprises of four encoder layers with a dropout ratio of 0.3. The output of the Transformer encoder is then directed towards a fully connected layer to resize the final output size, which is tailored to fit various tasks. 
In the training process, the parameters of MAE encoder and audio feature extractor are fixed without training. During the computation of loss, we flatten the sequence result of a clip and use the same loss function as Equ.~\ref{eq:1}, \ref{eq:2}, \ref{eq:3}.

%\subsection{Post-processing}
Since the final prediction is made frame-by-frame, applying a smoothing policy can enhance the stability of the predictions. 
We find that the cropped face images from videos occasionally miss some frames potentially due to the frames not being able to detect faces. 
Therefore, to maintain continuity, we use the nearest frame to replace the lost frame and produce the prediction for each frame.
Then, we use the Gaussian filter to smooth the probability for AU or EXPR and VA values. The sigma of Gaussian filter is set according to different tasks. The detailed setting can be seen in the experiment section. 

%% file: 4_Experiments.tex
\section{Experiment}

%%%%%%%%%%%%%%%%
\subsection{Experimental Setting}

We pre-processed all videos in the Aff-Wild2 datasets into frames by OpenCV and employ the OpenFace~\cite{baltrusaitis2018openface} detector to crop all facial images into $224\times 224$ scale. We find that OpenFace incorrectly detects some images and there are no faces in them. 
We use the NetEase Fuxi Youling crowdsourcing platform to check and remove incorrect images.
We pre-train MAE on large-scale facial image datasets for 800 epochs using the AdamW~\cite{loshchilov2017adamw} optimizer. We set the batch size as 4096 and the learning rate as 0.0024. Our training process is implemented based on PyTorch and trained on 8 NVIDIA A30 GPUs. 
In the MAE fine-tuning phase, we set the batch size to 512 and the learning rate to 0.0001, while using AdamW as the optimizer.
During TMF training, we set the video clip length to 100, with a batch size of 32 and a learning rate of 0.0001. The TMF training process was completed in approximately 20 epochs using the AdamW optimizer. All the experiments are carried out with the Fuxi Youling Platform that is based Agent-Oriented Programming (AOP) in order to facilitate task modeling.
% In the DCL training, we set the $\alpha$ to follow the distribution of $Beta(2,2)$, other experimental settings are the same as single BLB training.

%Moreover, we employed several training techniques to improve the model's performance. Specifically, we implemented a cosine annealing scheduler with a warmup to adjust the learning rate. We also utilized exponential moving average~(EMA) to stabilize the model weights during training and achieve a more robust result. Lastly, we leveraged the model soup approach~\cite{wortsman2022model} to further enhance the model's performance on the validation set. All the experiments are carried out with the Fuxi Youling Platform that is based Agent-Oriented Programming (AOP) in order to facilitate task modeling.

%%%%%%%%%%%%%%%%
\textbf{Metrics.} For AU detection and expression classification, we calculate the F1-Score~(F1) for each class to evaluate the prediction results. For VA estimation, we calculate the Concordance Correlation Coefficient (CCC) for valence and arousal respectively. The definition of CCC can be seen equ.~\ref{CCC}. In the case of ERI estimation, we utilize Pearson's Correlation Coefficient (PCC) for each class as the metric. The specific definitions for each challenge are as follows:
\begin{equation}
S_{AU} = \frac{1}{N_{au}}\sum_{}^{}  F1_{au_{i}} 
\label{equ:s_au}
\end{equation}

\begin{equation}
S_{EXP}= \frac{1}{N_{exp}}\sum_{}^{}  F1_{exp_{i}}
\label{equ:s_exp}
\end{equation}

\begin{equation}
S_{VA} = 0.5*(CCC(\hat{v},v)+CCC(\hat{a},a))
\end{equation}

\begin{equation}
S_{ERI} = \frac{1}{N_{exp}}\sum_{}^{} PCC(\hat{p}_{exp_{i}},p_{exp_{i}})
\end{equation}

\begin{equation}
    PCC = \frac{Cov({x},{\hat{x}})}{\delta _{{x}}\delta _{{\hat{x}}}}
    \label{equ:pcc}
\end{equation}
where $Cov(,)$ represents the covariance.

%%%%%%%%%%%%%%%%
\subsection{Results on Validation set}

\begin{table*}[!t]
\renewcommand\arraystretch{1.05} 
\centering
{
  \begin{tabular}{l|cccccccccccc|c}
    \hline
    \textbf{Val Set} &\textbf{AU1} &\textbf{AU2} &\textbf{AU4} &\textbf{AU6} &\textbf{AU7} &\textbf{AU10} &\textbf{AU12} &\textbf{AU15} &\textbf{AU23} &\textbf{AU24} &\textbf{AU25} &\textbf{AU26} &\textbf{Avg.}\\[1pt]
    \hline \hline
    Official & 55.26 & 51.35 & 56.70 & 67.25 & 75.75 & 75.11 & 75.82 & 31.21 & 17.34 & \textbf{33.77} & 83.90 & 42.26 & 55.86\\
    fold-1   & \textbf{64.94} & 52.31 & \textbf{71.84} & 74.59 & 69.10 & 69.92 & 74.05 & 35.38 & \textbf{28.51} & 21.56 & 77.97 & 42.14 & 56.86 \\
    fold-2   & 63.90 & \textbf{53.86} & 71.59 & \textbf{75.72} & \textbf{77.15} & \textbf{76.26} & \textbf{79.39} & 27.78 & 28.28 & 23.86 & \textbf{86.20} & 40.56 & \textbf{58.71}\\
    fold-3   & 61.80 & 51.10 & 60.85 & 70.70 & 73.08 & 75.15 & 74.84 & 29.79 & 23.71 & 27.35 & 79.22 & \textbf{48.33} & 56.33 \\
    fold-4   & 51.59 & 39.27 & 53.40 & 64.69 & 66.73 & 71.83 & 72.80 & \textbf{44.44} & 23.87 & 27.60 & 76.70 & 39.12 & 52.67 \\
    fold-5   & 52.69 & 44.59 & 60.90 & 70.67 & 70.11 & 73.25 & 76.04 & 43.84 & 23.86 & 10.30 & 76.96 & 39.96 & 53.60 \\
    \hline
  \end{tabular}}
  \caption{The AU F1 scores~(in \%) of models that are trained and tested on different folds (including the original training/validation set of \textit{Aff-Wild2} dataset). The highest and lowest scores are both indicated in bold.}
  \label{tab:AU_F1_val}
  %}
\end{table*}

\begin{table*}[!t]
\renewcommand\arraystretch{1.05} 
\centering
{  
  \begin{tabular}{l|cccccccc|c}
    \hline
    \textbf{Val Set} &\textbf{Neutral} &\textbf{Anger} &\textbf{Disgust} &\textbf{Fear} &\textbf{Happiness} &\textbf{Sadness} &\textbf{Surprise} &\textbf{Other} &\textbf{Avg.}\\[1pt]
    \hline \hline
    Official&  65.11 & 44.61 & 48.91 & 18.85 & 56.46 & 60.95 & 32.20 & 64.32 & 48.93 \\
    fold-1 &   61.92 & 34.22 & 46.96 & 16.96 & 53.65 & 77.77 & 30.08 & 50.12 & 46.46 \\
    fold-2 &   74.60 & 16.22 & 41.49 & \textbf{70.22} & \textbf{65.26} & 57.42 & \textbf{40.29} & 60.31 & 53.23 \\
    fold-3  &  62.16 & 36.00 & 21.99 & 19.40 & 60.12 & 65.68 & 32.38 & \textbf{72.16} & 46.24 \\
    fold-4 &   63.67 & \textbf{57.74} & \textbf{56.09} & 15.09 & 64.97 & \textbf{88.98} & 25.18 & 72.01 & \textbf{55.47} \\
    fold-5 &   \textbf{77.54} & 25.86 & 13.12 & 19.50 & 57.36 & 42.75 & 29.51 & 50.57 & 39.53 \\
    
    \hline
  \end{tabular}}
  \caption{The expression F1 scores~(in \%) of models that are trained and tested on different folds (including the original training/validation set of \textit{Aff-Wild2} dataset). The highest and lowest scores are both indicated in bold.}
  \label{tab:exp_F1_val}
  %}
\end{table*}

\textbf{Action Unit Detection Challenge.}
For AU detection, We evaluate the performance of our model using the F1 scores in equ.~\ref{equ:s_au}. To improve the generalization capability of our model, we also perform 5-fold cross-validation based on randomly split videos from the existing labeled data. We present the F1 scores of each AU and their average F1 in Tab.~\ref{tab:AU_F1_val}. From the table, we find that the performances of AU4, AU15, and AU24 in different data splits have a large variance~(AU4: 53.40\% $ \sim $ 71.84\%, AU15: 27.78\% $ \sim $ 44.44\%, AU24: 17.34\% $ \sim $ 28.51\%). The reason for this could be that the number of these AUs is relatively low, which means the learning for these AUs is not sufficient and heavily relies on the data splits.   

% \begin{table}[]
% \small
%     \centering
%     \setlength{\tabcolsep}{1.0mm}{
%     \begin{tabular}{c|cccccc}
%     \hline
%        \multirow{2}{*}{Method} & \multicolumn{6}{c}{Validation Set} \\ \cline{2-7} 
%         & Official & fold1 & fold2 & fold3 & fold4 & fold5 \\
%        \hline\hline
%        Our-MAE  & \textbf{0.5527} & 0.5430 & 0.5724 & 0.5440 & 0.5179  &  \textbf{0.5416} \\
%        Our-BLB & 0.5501 & \textbf{0.5547} & \textbf{0.5842} & \textbf{0.5441} & \textbf{0.5240} &  0.5337 \\
%        % Our-BLB & 0.5501 & 0.5547 & 0.5842 & 0.5441 & 0.5240 &  0.5337 \\
%        % Our-DCL & \textbf{0.5667} & \textbf{0.5647} & \textbf{0.5929} & \textbf{0.5460} & \textbf{0.5345} & 0.5411 \\
%        \hline
%     \end{tabular}
%     \caption{AU: F1-score on the official and 5-fold validation set.}
%     \label{tab:val_AU}    
%     }
% \end{table}

\textbf{Expression Classification Challenge.}
For the EXPR challenge, we also perform 5-fold cross-validation based on randomly selected video clips from existing labeled data. We show our experimental results on the official and 5-fold validation sets in Tab. \ref{tab:exp_F1_val}.  We evaluate the model by the average F1 metric in equ.~\ref{equ:pcc}. From the table, we also find the model performances for Anger, Disgust, and Fear are relatively unstable~(Anger: 16.22\% $ \sim $ 57.74\%, Disgust: 13.12\% $ \sim $ 56.09\%, Fear: 15.09\% $ \sim $ 70.22\%). At the same time, these three categories have the lowest frequency of occurrence in the dataset, which leads to their poor performances.

% \begin{table}[]
% \small
%     \centering
%     \setlength{\tabcolsep}{1.0mm}{
%     \begin{tabular}{c|cccccc}
%     \hline
%        \multirow{2}{*}{Method} & \multicolumn{6}{c}{Validation Set} \\ \cline{2-7} 
%         & Official & fold1 & fold2 & fold3 & fold4 & fold5 \\
%        \hline\hline
%        Our-MAE  & 0.4679 & 0.4203 & 0.4709 & 0.4241 & 0.5066  & 0.3493 \\
%        Our-BLB & \textbf{0.4817} & \textbf{0.4646} & \textbf{0.5323} & \textbf{0.4624} & \textbf{0.5547} & \textbf{0.3953} \\
%        % Our-BLB & 0.4817 & 0.4646 & 0.5323 & 0.4624 & 0.5547 & 0.3953 \\
%        % Our-DCL & \textbf{0.4952} & \textbf{0.4758} & \textbf{0.5376} & \textbf{0.4634} & \textbf{0.5589} & \textbf{0.3981} \\
%        \hline
%     \end{tabular}
%     \caption{EXPR: F1-score on the official and 5-fold validation set.}
%     \label{tab:val_expr}    
%     }
% \end{table}

\textbf{Valence-Arousal Estimation Challenge.}
For VA estimation, We evaluate the model by the CCC of Valence and Arousal in equ.~\ref{CCC}. To enhance the model generalization, we also perform the 5-fold cross-validation according to random video split in the existing labeled data. The experimental results can be seen in Tab. \ref{tab:va_F1_val}.

\begin{table}[!t]
\renewcommand\arraystretch{1.05} 
\centering
{
  \begin{tabular}{l|cc|c}
    \hline
    \textbf{Val Set} &\textbf{Valence} &\textbf{Arousal} & \textbf{Avg.}\\[1pt]
    \hline \hline
    Official&  0.4643 & 0.6407 &  0.5525\\
    fold-1 &   0.5927 & 0.6542 &  0.6234\\
    fold-2 &   0.5647 & 0.6267 &  0.5957\\
    fold-3  &  0.5679 & 0.6959 &  0.6319\\
    fold-4 &   0.5567 & 0.6456 &  0.6011\\
    fold-5 &   \textbf{0.6478} & \textbf{0.7056} &  \textbf{0.6767}\\
    
    \hline
  \end{tabular}}
  \caption{The VA CCC scores of models that are trained and tested on different folds (including the original training/validation set of \textit{Aff-Wild2} dataset). The highest and lowest scores are both indicated in bold.}
  \label{tab:va_F1_val}
  %}
\end{table}

\begin{table}[]
\small
    \centering
    \setlength{\tabcolsep}{4.0mm}
    {
    \begin{tabular}{c|cc}
    \hline
       \multirow{2}{*}{Team} & \multicolumn{2}{c}{Test Set} \\ \cline{2-3} 
        & Rank & F1-score  \\
       \hline\hline
       PRL~\cite{ABAW5_AU5} & \#5 & 51.01  \\
       SZFaceU~\cite{ABAW5_AU4} & \#4 & 51.28  \\
       USTC-IAT-United~\cite{ABAW5_AU3} & \#3 & 51.44  \\
       SituTech~\cite{ABAW5_AU2} & \#2 & 54.22  \\
       Ours & \#1 & \textbf{55.49}  \\
       \hline
    \end{tabular}
    \caption{Final competition results ~(average F1 score in \%) on the AU test set of ABAW5.}
    \label{tab:AU_test_set}    
    }
\end{table}

\subsection{Results on Test Set}
\textbf{Action Unit Detection Challenge}
In ABAW5 competition, we need to predict the labels of the official test set. Our proposed method achieves an average F1 score of 55.49\% for AU detection, thereby winning first place in this track (See Tab.~\ref{tab:AU_test_set}). 
The second-place team SituTech and the fourth-place team SZFaceU in this track also
leverage MAE features. Similar to our approach, they use a large-scale facial image dataset for MAE pre-training and use it as the feature extractor. 
Apart from this, SituTech incorporates additional visual features from DenseNet~\cite{iandola2014densenet}, IResNet100~\cite{iresnet}, and MobileNet~\cite{howard2017mobilenets}, pre-trained on the expression recognition task based on AffectNet~\cite{AffectNet}, FER+~\cite{barsoum2016training}, and RAF-DB~\cite{RAF}. In contrast, we fine-tune MAE on Aff-Wild2 and incorporate it into our multi-modal information training framework.  
SZFaceU proposes a Spatial-Temporal Graph Learning module to construct a graph representation for spatial-Temporal features. 
The third-place team proposes to use LANet~\cite{LANet} to extract the local features for AU. They model the correlations between AUs through Graph Neural Networks.
PRL utilizes the Video Vision Transformer~\cite{arnab2021vivit} to capture the temporal expression movements in the video.

\textbf{Expression Classification Challenge}
In Tab.~\ref{tab: EXP_test_set}, our approach achieves an average F1 score of 41.21\% in EXPR track, winning first place in this track. Our approach, as well as the second-place team (SituTech), the third-place team (CtyunAI) incorporate multi-modal information from both vision and acoustic modalities. 
Besides the above-mentioned vision features, SituTech uses the audio features of Hubert~\cite{hubert}, wav2vec~\cite{baevski2020wav2vec} and ECAPA-TDNN~\cite{ecapa}.
CtyunAI use three kinds of vision features: arcface~\cite{deng2019arcface}, EfficientNet-b2~\cite{savchenko2022video} and DAN~\cite{wen2021dan} and two kinds of audio features: wav2vec2~\cite{baevski2020wav2vec} and wav2vec2-emotion~\cite{pepino2021wav2vec2_emo}. 
HFUT-MAC uses POSTER2~\cite{mao2023poster} as the feature extractor and employs an affine module to align features of varying dimensions to a uniform dimension. Subsequently, a transformer is utilized to integrate the temporal features. 
HSE-NN-SberAI applies EfficientNet CNN from HSEmotion library~\cite{savchenko2022hsemotion} to extract effective prior expression knowledge and use MLP for classification.

\begin{table}[]
\small
    \centering
    \setlength{\tabcolsep}{4.0mm}
    {
    \begin{tabular}{c|cc}
    \hline
       \multirow{2}{*}{Team} & \multicolumn{2}{c}{Test Set} \\ \cline{2-3} 
        & Rank & F1-score  \\
       \hline\hline
       HSE-NN-SberAI~\cite{ABAW5_EXPR5} & \#5 & 32.92  \\
       HFUT-MAC~\cite{ABAW5_EXPR4} & \#4 & 33.37  \\
       CtyunAI~\cite{ABAW5_EXPR3} & \#3 & 35.32  \\
       SituTech~\cite{ABAW5_AU2} & \#2 & 40.72  \\
       Ours & \#1 & \textbf{41.21}  \\
       \hline
    \end{tabular}
    \caption{Final competition results ~(average F1 score in \%) on the EXPR test set of ABAW5.}
    \label{tab: EXP_test_set}    
    }
\end{table}

\textbf{VA Estimation Challenge}
In Table~\ref{tab: VA_test_set}, our approach achieves an average CCC score of 0.6372 in the VA track, securing second place. Notably, our results are highly competitive with the first-place team's score of 0.6414. 
After extracting three modalities of information: vision, acoustic and text, CBCR use TCN to capture the temporal features and channel attention network (CAN) for features fusion.
The methods of other teams are roughly similar to the analysis of other challenges.

\begin{table}[]
\small
    \centering
    \setlength{\tabcolsep}{4.0mm}
    {
    \begin{tabular}{c|cc}
    \hline
       \multirow{2}{*}{Team} & \multicolumn{2}{c}{Test Set} \\ \cline{2-3} 
        & Rank & CCC  \\
       \hline\hline
       HFUT-MAC~\cite{ABAW5_EXPR4} & \#5 & 0.5342  \\
       CtyunAI~\cite{ABAW5_EXPR3} & \#4 & 0.5666  \\
       CBCR~\cite{ABAW5_VA3} & \#3 & 0.5913  \\
       SituTech~\cite{ABAW5_AU2} & \#1 & \textbf{0.6414}  \\
       Ours & \#2 & 0.6372  \\
       \hline
    \end{tabular}
    \caption{Final competition results ~(average CCC) on the VA test set of ABAW5.}
    \label{tab: VA_test_set}    
    }
\end{table}

\subsection{Ablation Studies}
In this section, we conduct several experiments to discuss the significance of each module of our approach, including MAE finetuning training, Temporal and Multi-modal Fusion~(TMF), the selection of features and the policies of post-processing. All the experiments are perform on the official training and validation set.

\begin{table}[]
\small
    \centering
    \setlength{\tabcolsep}{1.0mm}
    {
    \begin{tabular}{c|c|c|c}
    \hline
        Method & AU  & EXPR  & VA \\ 
       \hline\hline
       w/o. MAE finetune & 51.04 & 41.34 & 0.5169\\
       w/o TMF & 55.27 & 46.79 & 0.5483\\
       Ours & \textbf{55.86} & \textbf{48.93} & \textbf{0.5525}\\
       \hline
    \end{tabular}
    \caption{Ablation studies that discuss the significance of MAE fine-tune and Temporal and Multi-modal Fusion.  The evaluation metrics used for AU, EXPR, and VA are average F1 (\%), average F1 (\%), and CCC, respectively. }
    \label{tab: abla1}    
    }
\end{table}

\textbf{MAE finetuning training.}
To prove the effectiveness of our procedure of finetuning MAE on Aff-wild2~(See Fig.~\ref{fig:pipeline} (left: step 2)), we conduct the experiment that directly uses the initial MAE features to operate the TMF training. The result can be seen in Tab.~\ref{tab: abla1} (w/o. MAE finetune). It can be found that there is a significant drop in the performance metrics across all three tracks. The average F1 score for AU decreases from 55.86\% to 51.04\%, and for EXPR it decreased from 48.93\% to 41.34\%. The average CCC of VA also decreases from 0.5525 to 0.5169. This illustrates that MAE finetuning can effectively exploit the static vision features based on a single image, which provides valuable prior knowledge for learning the temporal vision feature.

\textbf{Temporal and Multi-modal Fusion.}
To illustrate the significance of TMF, we conduct an experiment that removes the TMF and uses the model trained on the single frame for evaluation. The results are presented in Tab.~\ref{tab: abla1}~(w/o TMF). It is evident that the inclusion of TMF leads to an  improvement in AU F1, EXPR F1, and VA CCC scores by 0.59\%, 2.14\% and 0.0042, respectively. This proves that the temporal and multi-modal features can further exploit more hidden clues for these three tasks. Notably, TMF is particularly effective for EXPR compared to other challenges.

\textbf{The choice of multi-modal features.}
In this section, we analyze the impact of using different multi-modal features, as presented in Tab.~\ref{tab: abla2}. Specifically, we experiment with two types of vision features: expression embedding from DLN~\cite{DLN} and the aforementioned MAE features. Our results demonstrated that using MAE features alone has a more significant advantage in all three challenges. Hence, for multi-modal information fusion, we only utilized the features extracted by MAE.
In terms of audio features, we try three kinds of features: Hubert~\cite{hubert}, wav2vec2~\cite{baevski2020wav2vec} and Vggish~\cite{vggish}. From Tab.~\ref{tab: abla2}, we observe that the  utilization of audio features does not enhance the performance of the AU task. However, the incorporation of audio features enhances the performance of the EXPR and VA challenges. 
For EXPR and VA tracks, our experiments show that the best performance is achieved by combining Hubert and Vggish features.

\begin{table}[]
\small
    \centering
    \setlength{\tabcolsep}{0.8mm}
    {
    \begin{tabular}{c|c|c|c|c}
    \hline
        Vis\_fea & Audio\_fea & AU  & EXPR  & VA \\ 
       \hline\hline
        MAE & - & \textbf{55.86} & 47.73 & 0.5483\\
        DLN & - & 50.49 & 39.12 & 0.4789 \\
        MAE + DLN & - & 54.76 & 44.69  & 0.5169\\
       \hline\hline
       MAE & Hubert  & 54.89 &  48.03 & 0.5405\\ 
       MAE & Wav2vec2 & 53.14 &  47.91  & 0.5364\\
       MAE & Vggish  & 51.68 &  47.71 & 0.5157\\
       MAE & Hubert+Vggish & 52.28 & \textbf{48.93} & \textbf{0.5525} \\
       MAE & Hubert+wav2vec2 & 52.55 & 48.18  & 0.5471 \\
       MAE & Hubert+wav2vec2+Vggish  & 53.54 & 47.89 & 0.5521\\
    \hline
    \end{tabular}
    \caption{Ablation studies that discuss the significance of different selection of vision and audio features. The evaluation metrics used for AU, EXPR, and VA are average F1 (\%), average F1 (\%), and CCC, respectively.}
    \label{tab: abla2}    
    }
\end{table}

%\subsubsection{Post-processing policies}
%\noindent\textbf{Smooth Policy}
\textbf{Smooth Policy}
Due to the frame-by-frame prediction of the challenges, smooth policies can effectively eliminate some noisy predictions and enhance prediction stability. In this part, we discuss the different policies we used for smoothing in Tab.~\ref{tab:abla3}. From the table, we can find that the metrics have varying degrees of improvement by using smooth policies. We use Scipy library to realize these smooth approaches. In different challenges, there exist slight differences in hyper-parameter settings. For instance, we set the sliding window size of median\_filter and average\_filter in AU, EXP and VA tracks as 10, 25 and 50, respectively. We set the sigma of gaussian\_filter in AU, EXP and VA tracks as 5, 25 and 25, respectively. This is because human facial physical movements of expressions tend to change frequently over a short period of time, whereas emotional states often exhibit a small range of change. Therefore, AU changes are more sensitive compared to EXP and VA. Our final prediction for the official test set combines some of the mentioned smooth policies. 

\begin{table}[]
\small
    \centering
    \setlength{\tabcolsep}{0.8mm}
    {
    \begin{tabular}{c|c|c|c}
    \hline
        Policy & AU &  EXP  &  VA \\ 
       \hline\hline
        w/o. smooth & 55.86 & 48.93 & 0.5525 \\
        gaussian\_filter & \textbf{56.01} & \textbf{49.16} & \textbf{0.5640} \\
        median\_filter & 55.91 & 48.98 & 0.5537 \\
        average\_filter  & 55.95 & 49.08 & 0.5595 \\
    \hline
    \end{tabular}
    \caption{Ablation studies that discuss the influence of different smooth policies. The evaluation metrics used for AU, EXPR, and VA are average F1 (\%), average F1 (\%), and CCC, respectively.}
    \label{tab:abla3}    
    }
\end{table}

%% file: 5_Conclusion.tex
\section{Conclusion}
This paper introduces our submission to the ABAW5 competition for the tasks of AU detection, expression recognition, and VA estimation. Our approach begins by pre-training a MAE in a self-supervised manner, using a large-scale facial image dataset. This enables the MAE to learn a variety of general features associated with human faces. Subsequently, we finetune the MAE using static images from Aff-wild2 dataset. Then we propose Temporal and Multi-modal Fusion~(TMF) to exploit the multi-modal information from the vision and audio temporal features. In participating in the ABAW5 competition, we won the first prizes in the AU track and EXPR track and second prize in the VA track. The quantities ablation studies indicate that each module and procedure of our approach can improve the model performance for affective tasks.

\section*{Acknowledgments}
The experiments and the data management and storage are supported by Netease Fuxi  Youling platform, based on Fuxi Agent-Oriented Programming~(AOP) system that is carefully designed to facilitate task modeling. This work is also supported by the 2022 Hangzhou Key Science and Technology Innovation Program (No. 2022AIZD0054), and the Key Research and Development Program of Zhejiang Province (No. 2022C01011). 